%% file: main.tex
\documentclass{article}

\usepackage[preprint]{neurips_2026}
\usepackage{amssymb}
\usepackage{amsmath}
\usepackage{amsfonts}
\usepackage{bbm}
\usepackage{bm}
\usepackage{xcolor} 
\usepackage{adjustbox}
\usepackage{times}
\usepackage{latexsym}
\usepackage{booktabs}
\usepackage{multirow}
\usepackage[T1]{fontenc}
\usepackage[utf8]{inputenc}
\usepackage{microtype}
\usepackage{inconsolata}
\usepackage{graphicx}
\usepackage{comment}
\usepackage{adjustbox}
\usepackage{algorithm}
\usepackage{algpseudocode}
\usepackage{colortbl}
\usepackage{nicefrac}
\usepackage{wrapfig}
\usepackage{hyperref}           
\hypersetup{
    colorlinks=true,
    linkcolor=blue,
    filecolor=magenta,      
    urlcolor=cyan,
    citecolor=blue,
}
\usepackage{cleveref}
\usepackage{url}

\newcommand\Tensor{\mathcal}

\title{Strategic Over-Parameterization for Generalizable Low-Rank Adaptation}

\author{%
  \normalfont\normalsize
  \begin{tabular}{c}
    \textbf{Jing Gao}$^{1,2,3}$,~~\textbf{Zhong-Yi Lu}$^{4}$,~~\textbf{Pan Zhang}$^{1,2,3}$,~~\textbf{Ze-Feng Gao}$^{4}$\textsuperscript{$\dagger$} \\[0.4em]
    \mdseries
    $^{1}$School of Fundamental Physics and Mathematical Sciences,\\
    Hangzhou Institute for Advanced Study, UCAS, Hangzhou 310024, China\\
    $^{2}$School of Physical Sciences, University of Chinese Academy of Sciences,\\
    No.~19A Yuquan Road, Beijing 100049, China\\
    $^{3}$CAS Key Laboratory of Theoretical Physics, Institute of Theoretical Physics,\\
    Chinese Academy of Sciences, Beijing 100190, China\\
    $^{4}$School of Physics and Key Laboratory of Quantum State Construction\\
    and Manipulation (Ministry of Education), Renmin University of China,\\
    Beijing 100872, China\\[0.3em]
    \texttt{gaojing23@mails.ucas.ac.cn}, \quad \texttt{zfgao@ruc.edu.cn}
  \end{tabular}%
}

\begin{document}
\maketitle

\input{section/sec-abstract}
\input{section/sec-introduction}
\input{section/sec-relatedwork}

\input{section/sec-preliminary}
\input{section/sec-method}
\input{section/sec-experiment}
\input{section/sec-conclusion}

\bibliographystyle{abbrvnat}
\bibliography{custom.bib}
\newpage
\appendix
\input{section/sec-appendix}


\end{document}

%% file: section/sec-abstract.tex
\begin{abstract}

Adapting large language models (LLMs) to downstream tasks via full fine-tuning is increasingly impractical due to its computational and memory demands. Parameter-efficient fine-tuning (PEFT) approaches such as Low-Rank Adaptation (LoRA) mitigate this by confining updates to a compact set of trainable parameters, but this aggressive reduction often sacrifices generalization, especially under transfer across heterogeneous tasks and domains. We revisit the tension between parameter efficiency and adaptation capacity, and ask whether the two are truly at odds. We answer in the negative by introducing LoRA-Over, a framework grounded in a simple principle: enrich the optimization landscape during training, then collapse the enrichment at inference. LoRA-Over injects auxiliary parameters into the low-rank adapters during training to broaden the effective hypothesis space, and through a decomposition-based reformulation folds them back into a standard low-rank structure with negligible reconstruction error, keeping inference cost identical to vanilla LoRA. Since not all weight matrices benefit equally from added capacity, we further propose two scheduling strategies, one statically predefined and one dynamically determined at runtime, that direct extra capacity where most needed. We evaluate LoRA-Over on language understanding (GLUE, T5-Base), dialogue (MT-Bench), arithmetic reasoning (GSM8K), and code generation (HumanEval), using LLaMA 2-7B and LLaMA 3.1-8B. Across all benchmarks and scales, LoRA-Over consistently outperforms vanilla LoRA, showing that principled over-parameterization designed to vanish at inference is an effective lever for improving PEFT generalization. Code will be released upon acceptance.
\end{abstract}

%% file: section/sec-introduction.tex
\section{Introduction}
\label{sec-intro}
The advent of modern large language models (LLMs) has fundamentally revolutionized the field of machine learning, achieving exceptional performance across a wide range of tasks by leveraging massive pre-training on trillion-token corpora \citep{devlin2019bert,liu2019roberta,radford2019language,brown2020language}. A dominant strategy for adapting pre-trained LLMs to downstream applications is full-parameter fine-tuning (commonly referred to as full fine-tuning) \citep{qiu2020pre,raffel2020exploring}, wherein all parameters of the model are updated using gradient-based optimization. While this approach delivers outstanding results, it presents significant challenges, particularly in terms of substantial storage requirements and high computational costs, making its practical deployment challenging.

To overcome these limitations, parameter-efficient fine-tuning (PEFT) methods have emerged as a promising solution. PEFT techniques, such as Low-Rank Adaptation (LoRA) \citep{hu2022lora}, aim to efficiently adapt large language models to specific domains by freezing the majority of pre-trained parameters and updating only a small subset of them. 
This approach alleviates the computational complexity and memory overhead associated with full fine-tuning while maintaining a reasonable level of performance. However, the drastic reduction in trainable parameters can compromise the model's ability to generalize, often resulting in suboptimal fine-tuning outcomes compared to full fine-tuning.

Motivated by this performance gap, we propose strategically over-parameterizing low-rank matrices during fine-tuning to enhance generalization. By utilizing matrix decomposition \citep{henry19928,tucker1966some,oseledets2011tensor, gao2020compressing}, we temporarily expand the parameter space during training and re-collapse it for inference, ensuring zero architectural overhead. However, this approach faces the dual challenges of performance degradation resulting from the accumulation of decomposition inaccuracies across deep Transformer layers and the computational redundancy caused by indiscriminately over-parameterizing all matrices regardless of their unequal contributions to task performance \citep{zhang2022platon,voita2019analyzing}. Therefore, identifying and selectively over-parameterizing the most critical matrices is essential.

To address these dilemmas, we employ the Matrix Product Operator (MPO) \citep{pirvu2010matrix} from quantum many-body physics as our primary method for matrix decomposition, which enables near-lossless matrix reconstruction \citep{gao2020compressing}, making it ideal for high-fidelity over-parameterization. Building on this, we introduce \textbf{LoRA-Over} (Over-Parameterization for Low-Rank Adaptation), a framework that adaptively identifies critical matrices through two distinct mechanisms: a predefined strategy that estimates importance via loss sensitivity \citep{voita2019analyzing}, and a runtime strategy that uses gradient variations \citep{hou2020dynabert} as a dynamic proxy. By selectively expanding key parameters, LoRA-Over enhances fine-tuning performance across diverse downstream tasks while remaining task-agnostic and inference-efficient.

We conduct extensive experiments to evaluate the efficacy of LoRA-Over. For natural language understanding tasks, we assess the performance of T5-Base \citep{raffel2020exploring} on the General Language Understanding Evaluation (GLUE) subset \citep{wang2018glue}. For dialogue generation tasks, mathematical reasoning tasks, and code generation tasks, we apply our method to Llama 2-7B \citep{touvron2023llama} and Llama 3.1-8B \citep{grattafiori2024llama}, evaluating their performance on the MT-Bench dataset \citep{zheng2023judging}, the GSM8K dataset \citep{cobbe2021training}, and the HumanEval dataset \citep{chen2021evaluating}, respectively.
Experimental results demonstrate that LoRA-Over significantly outperforms vanilla LoRA, achieving substantial performance gains. For instance, LoRA-Over consistently outperforms vanilla LoRA by 5.95\% on the GLUE subset with T5-Base, and by 0.31, 12.13\%, 5.20\% on MT-Bench, GSM8K, and HumanEval with Llama 2-7B, respectively. Additionally, LoRA-Over also consistently outperforms vanilla LoRA by 0.11, 6.17\%, 1.75\% on MT-Bench, GSM8K, and HumanEval with Llama 3.1-8B, respectively. These results underscore the versatility and effectiveness of LoRA-Over in addressing the limitations of vanilla LoRA while preserving computational efficiency. 

%% file: section/sec-relatedwork.tex
\section{Related work}
\label{sec-related-work}

\paragraph{Over-parameterization in learning process.}
Over-parameterization has demonstrated its effectiveness in multiple dimensions of deep learning. Previous studies have highlighted its utility in improving model initialization \citep{arpit2019benefits}, optimizing training dynamics through improved convergence \citep{allen2019convergence,gao2021global,du2018gradient}, and enhancing generalization capabilities \citep{allen2019learning}. Sparked by the hypothesis of lottery theory \citep{frankle2018lottery}, subsequent research has further emphasized its potential to increase training efficiency \citep{malach2020proving,pensia2020optimal} and improve model performance \citep{chen2020lottery,brix2020successfully,prasanna2020bert}. In particular, in-time over-parameterization strategies \citep{liu2021we} have been employed to bridge the performance gap between sparse and dense network training regimes. Building on these foundations, we propose to leverage over-parameterization as a principled approach to unlock the latent potential of the LoRA method, boosting its fine-tuning performance.

\paragraph{Tensor decomposition in neural network.}
Tensor decomposition has emerged as a cornerstone technique for improving the efficiency of neural network training and inference. Its versatility is evident in applications including model compression \citep{gao2020compressing}, lightweight fine-tuning \citep{liu2021enabling,gao2023small}, and knowledge distillation \citep{zhan2024over}. Pioneering works have extensively utilized these methods to decompose parameter matrices, achieving significant compression ratios for linear layers \citep{novikov2015tensorizing} and convolutional kernels \citep{garipov2016ultimate}. Beyond model compression, recent advances have further showcased their adaptability: MPO-based decomposition efficiently scales the MoE framework for dynamic model capacity \citep{gao2022parameter}, while parameterized tensor formats enable resource-conscious fine-tuning of ALBERT \citep{liu2021enabling}. Diverging from conventional paradigms that prioritize dimensionality reduction, our work exploits tensor decomposition from an inverse perspective: we deploy it as a latent space bridge to implicitly over-parameterize adapters by mapping low-dimensional parameters to high-dimensional latent spaces during fine-tuning.

\paragraph{Variants of LoRA in LLMs.}
LoRA is a widely used technique for fine-tuning LLMs, significantly reducing their resource requirements. Given its efficiency, numerous variants have been developed to improve the original method. One major line of research has focused on optimizing initialization and dynamic rank allocation. For instance, PiSSA \citep{meng2024pissa} initializes adapters using principal singular components via SVD, contrasting with standard random initialization. Meanwhile, methods like rsLoRA \citep{kalajdzievski2023rank} and AdaLoRA \citep{zhang2023adalora} dynamically adjust rank, scaling coefficients, and parameter budgets based on real-time data distribution and parameter importance. Furthermore, DoRA \citep{liu2024dora} decouples the magnitude and direction of pre-trained weights, while LoRA+ \citep{hayou2024lora+} introduces distinct learning rates to mitigate sub-optimal training dynamics. However, despite these advancements, these methods inherently maintain the shallow two-matrix ($A \times B$) multiplication structure in fine-tuning, which restricts the expressiveness and optimization trajectory of the adaptation modules. Another relevant trajectory has explored higher-order tensor operations for PEFT. Works such as LoHA \citep{hyeon2021fedpara} and LoKr \citep{edalati2025krona} employ Hadamard and Kronecker products, respectively, to restructure weight updates. Similarly, TT-LoRA \citep{anjum2024tensor} utilizes Tensor-Train to further reduce the parameter count of adaptation weights. Crucially, these approaches predominantly treat tensor networks as a compression mechanism. While highly parameter-efficient, they often encounter optimization bottlenecks and inevitably compromise downstream performance compared to vanilla LoRA. In contrast, we leverage MPO-based over-parameterization to expand the optimization space during training, rather than for compression. This transient structure enhances expressiveness but contracts back into original matrices for inference, achieving superior performance with zero deployment overhead.

%% file: section/sec-preliminary.tex
\section{Preliminary}
\paragraph{Tensor.} 

We denote a tensor $\Tensor{T}_{i_1,i_2,\dots, i_m}$ as an array with $m$ indices, where $\{i_1, i_2,\dots, i_m\}$ denotes the dimensions of the $m$ indices, respectively. The intrinsic tensor-based nature of both data and trainable parameters in deep learning establishes tensor representations as fundamental building blocks in neural networks.

\paragraph{Tensor product.}
As a fundamental construct in linear algebra, the tensor product formalism serves as a cornerstone in quantum mechanical analysis and also remains indispensable for both conceptual advances and computational protocols in many-body physics. Considering $\{\psi_i\}_{i=1}^{p}$ and $\{\phi_j\}_{j=1}^{q}$ are the orthonormal basis of tensors $\Tensor{T}^{(1)}$ and $\Tensor{T}^{(2)}$, respectively. The $\otimes$ denotes the tensor product. Formally, the tensor contraction of  $\Tensor{T}^{(1)} = \sum_{i=1}^{p} \alpha_i\psi_{i}$ and $\Tensor{T}^{(2)} = \sum_{j=1}^{q} \beta_j\phi_{j}$ is defined as follow:
\begin{equation}
    \Tensor{T}^{(1)} \otimes \Tensor{T}^{(2)} =  \sum_{i=1}^{p} \sum_{j=1}^{q}\alpha_i \beta_j \psi_{i}\otimes \phi_{j}.
\end{equation}

\paragraph{Tensor decomposition.}
Tensor decomposition can be regarded as the inverse operation of the tensor product. The SVD algorithm serves as a widely utilized mathematical framework for tensor decomposition. Given a tensor $\Tensor{T} \in \mathbb{R}^{i_1 \times \cdots \times i_n}$, through an iterative sequence of $n$ SVD operations, the original tensor can be factorized into $n$ hierarchically structured local tensors ${\{\Tensor{T}^{(k)}\}}_{k=1}^{n}$. In contrast, the decomposed tensors can reconstruct the original tensor $\Tensor{T}$ by sequentially utilizing the tensor product operator.

%% file: section/sec-method.tex
\section{Method}
\label{method}

\begin{figure*}
    \centering
    \includegraphics[width=.95\textwidth]{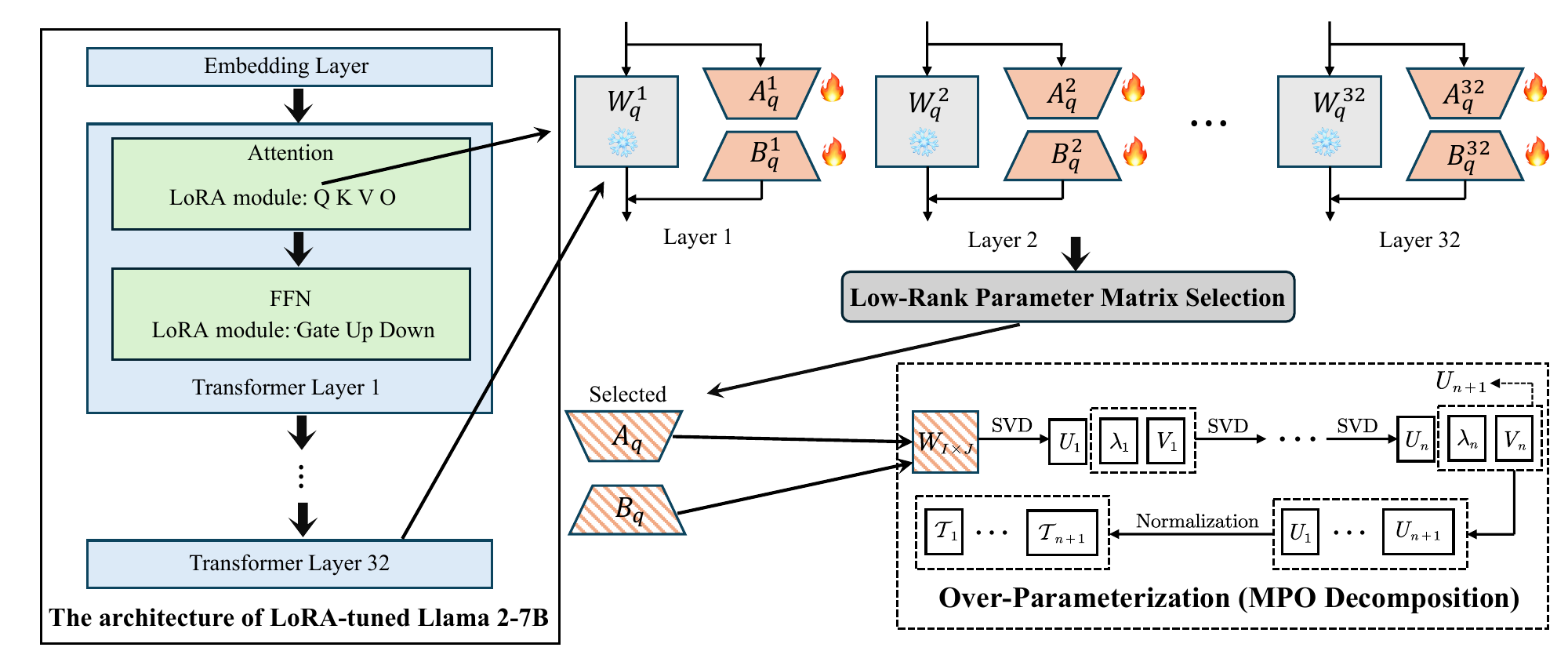}
    \caption{Workflow of LoRA-Over. Taking Llama-2-7B as an illustrative case, the top-$k$ highest-scoring low-rank matrices of the $Q$ modules are selected for over-parameterization.}
    \label{fig:main1}
\end{figure*}

\begin{wrapfigure}{r}{0.5\textwidth}
  \includegraphics[width=0.5\textwidth]{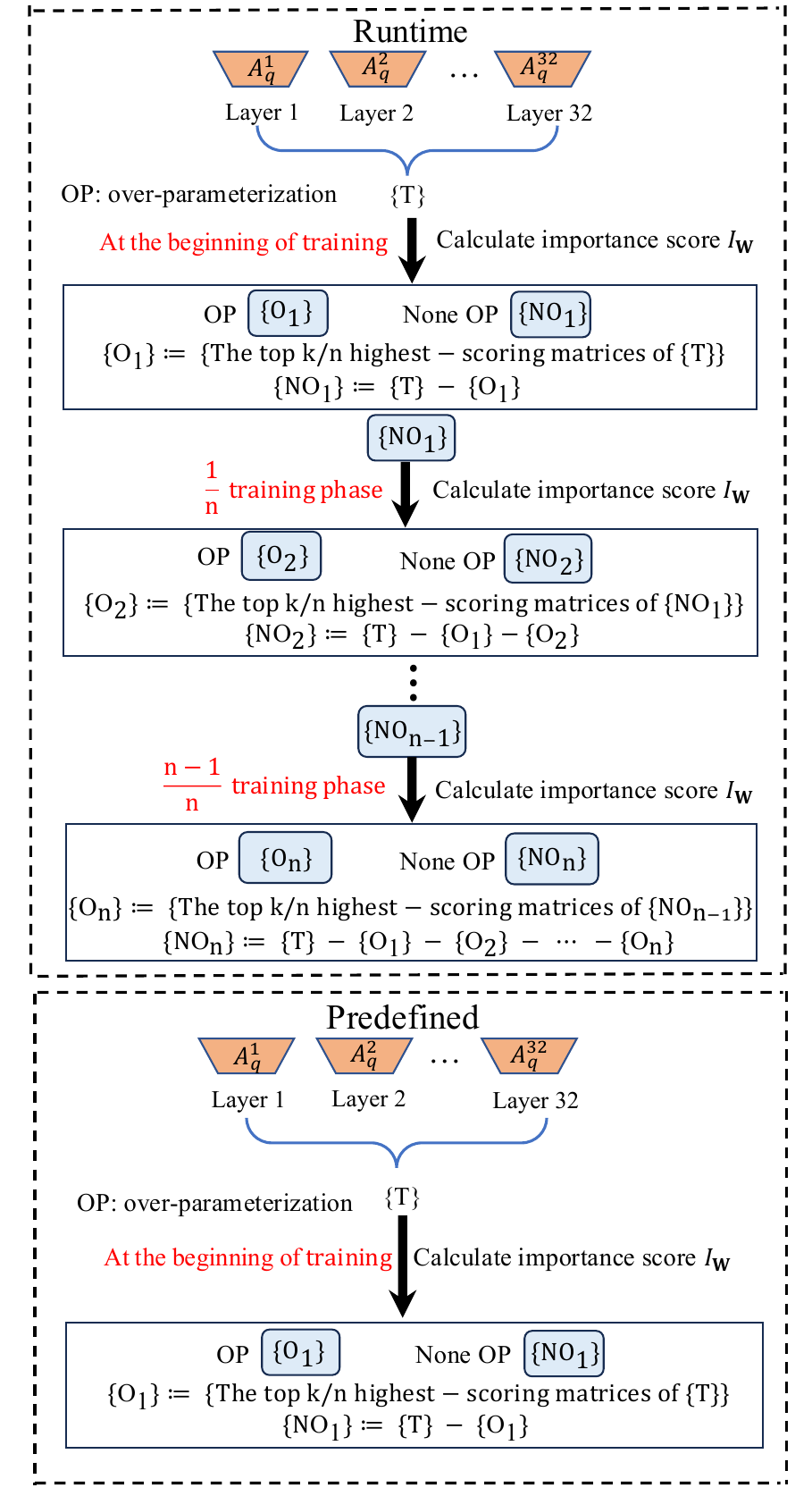}
  \caption{Overview of low-rank parameter matrix selection. We propose two selection strategies: runtime and predefined. To illustrate these processes, we use the $LoRA\_A$ matrices within the $Q$ module of Llama-2-7B as a representative case. In this setup, a top-$N$ of $k$, and a split number of $n$ are employed.
  }
  \label{fig:main2}
\end{wrapfigure}

In this section, we first give a review of vanilla LoRA. Subsequently, we propose our LoRA-Over, which utilizes MPO to increase the number of low-rank matrix parameters and introduces the over-parameterized low-rank parameter matrix selection strategies. The workflow of our approach is presented in \Cref{fig:main1}, and the overview of low-rank parameter matrix selection is provided in \Cref{fig:main2}. We also compare various tensor decomposition methods with MPO decomposition, and explicitly distinguish our approach from OPF \citep{gao2023small}, which can be found in \Cref{appendix:discussion}.

\subsection{Revisit the LoRA method}
LoRA \citep{hu2022lora} is a PEFT framework that introduces low-rank matrix $A \in\mathbb{R}^{r \times d_2}$ and $B\in \mathbb{R}^{d_1 \times r}$($r \ll min(d_1, d_2)$). During training, the pre-trained weight matrix $W_0$ remains frozen, while the model is updated by training the low-rank matrix $A$ and $B$, which extremely diminishes the number of trainable parameters, thus decreasing the computational cost of fine-tuning. In particular, the mathematical expression of LoRA is defined as: 
\begin{equation}
    W = W^{(0)} + \Delta W =W^{(0)} + \frac{\alpha}{r}BA,
\end{equation}
    where $\alpha$ and $r$ are hyper-parameters of scaling-factor and LoRA rank. Vanilla LoRA initializes $A$ with Kaiming normal distribution \citep{he2015delving}, while $B$ adopts zero initialization. This initialization strategy ensures $BA = 0$ at the beginning of training. Additionally, the low-rank matrix can be merged into the pre-trained matrix, and LoRA does not introduce any extra latency during the inference compared with full funetuning.

\subsection{Over-paramterization low-rank matrix via MPO decomposition}

To enhance the vanilla LoRA method by leveraging the benefits of over-parameterization during the fine-tuning process, the proposed approach utilizes the MPO, a tensor network decomposition technique, to expand the model's parameter space. Specifically, the methodology first details the fundamental principles of MPO decomposition. Subsequently, we describe the adaptation and application of this MPO framework to construct over-parameterized low-rank parameter matrix.

\paragraph{Matrix product operator decomposition.}
MPO decomposition efficiently factorizes a parameter matrix $\mathbf{W}\in \mathbb{R}^{I \times J}$ into a sequential product of multiple tensors \citep{gao2020compressing}. Formally, the MPO decomposition of a parameter matrix $\mathbf{W}\in \mathbb{R}^{I \times J}$ yields an ordered sequence of $m$ local tensors $\{\mathcal{T}^{(k)}\}^m_{k=1}$ can be denoted as:
\begin{equation}
    \mathrm{MPO}(\mathbf{W})=\bigotimes^m_{k=1} \mathcal{T}^{(k)}[d_{k-1},i_k,j_k,d_{k}], \label{mpo_define}
\end{equation}
where the tensor $\mathcal{T}^{(k)}[d_{k-1},i_k,j_k,d_{k}]$ is a 4th-order tensor with size $[d_{k-1},i_k,j_k,d_{k}]$, in which $\prod^m_{k=1} i_k=I,\prod^m_{k=1} j_k=J$, and $d_0=d_m=1$. The concept of a bond, introduced to link two sequence tensors, has been adopted following the work of \citep{pirvu2010matrix}.  The dimension of the bond $d_k$ is denoted as: 
\begin{equation}
    d_k=\mathrm{min}(\prod^k_{p=1}i_p \times j_p,\ \prod^n_{p=k+1}i_p \times j_p).\label{d_k}
\end{equation}
A deterministic mapping process from the parameter matrix $\mathbf{W}$ to multiple high-order tensors $\{\mathcal{T}^{(k)}\}^m_{k=1}$ is defined by the given tensor sizes $\{d_k\}^m_{k=1}$, $\{i_k\}^m_{k=1}$, and $\{j_k\}^m_{k=1}$. Through iterative matrix reshaping and SVD decomposition \citep{henry19928} executed on $m$-turns, the MPO process continuously reduces the size of the parameter matrix and sequentially generates decomposed tensors. Reshaping is applied during the $k$-th turn to the matrix of the previous turn $\mathbf{W}_{k-1}$, transforming it into the matrix $\mathbf{W}^{\prime}_{k-1}$ with the first dimension $d_{k-1}\times i_k \times j_k$. Subsequently, we decompose it via SVD as:
\begin{equation}
    \mathbf{U}\lambda \mathbf{V}^\mathrm{T}=\mathrm{SVD}(\mathbf{W}^{\prime}_{k-1}),
\end{equation}
where $\mathbf{U}$ and $\mathbf{V}$ are complex unitary matrix, $\lambda$ is a rectangular diagonal matrix with non-negative real numbers on the diagonal. Inspired by truncated SVD methods \citep{henry19928}, corresponding to the $d_k$ largest singular values, the first $d_k$ columns of $\mathbf{U}$ form the decomposed tensor $\mathcal{T}^{(k)}$, which is subsequently reshaped to match the dimensions of $[d_{k-1},i_k,j_k,d_{k}]$. Furthermore, $\lambda \mathbf{V}^{\mathrm{T}}$ is assigned as the output parameter matrix $\mathbf{W}_k$ for the decomposition in the subsequent turns. Following $m$-turn iterations, the decomposition results in a set of multiple high-order tensors $\{\mathcal{T}^{(k)}\}^m_{k=1}$. The original parameter matrix $\mathbf{W}$ can be recovered almost losslessly by contracting these tensors sequentially \citep{gao2020compressing}. You can find the comprehensive algorithm in \Cref{alg:mpo} of \Cref{appendix_algorithm}.

\paragraph{Over-parameterizing low-rank matrix.}
Utilizing the MPO methodology, we strategically amplify low-rank matrix parameterization scales during fine-tuning to exploit advantages inherent in structured over-parameterization. More precisely, the MPO method enables the decomposition of selected low-rank parameter matrix into multiple tensors according to \Cref{mpo_define}. The values of $\{d_k\}^m_{k=1}$, $\{i_k\}^m_{k=1}$, and $\{j_k\}^m_{k=1}$ govern the increase in parameter number in matrix W after MPO decomposition. The detailed added parameter number $N_{add}$ derives from the following calculation procedure:
\begin{equation}
    N_{a d d}=\sum_{k=1}^{m} i_{k} j_{k} d_{k-1} d_{k}-\prod_{k=1}^{m} i_{k} j_{k}.\label{add}
\end{equation}
Following the formalism of \Cref{d_k}, the determination of $\{d_k\}^m_{k=1}$ is based on $\{i_k;j_k\}^m_{k=1}$. Control over the number of added parameters is achieved by adjusting the values of $\{i_k;j_k\}^m_{k=1}$ within the MPO decomposition strategy. Consequently, the fine-tuning process allows the adoption of MPO on selected low-rank parameter matrix to generate the corresponding multiple tensors. This methodology enables scaling of the model's total parameter number, effectively enhancing its over-parameterization. After achieving convergence by fine-tuning the over-parameterized low-rank parameter matrix, tensor contraction is performed on the decomposed tensors to reconstruct the model's parameter matrices. The resulting model preserves identical parameter number and inference latency to the original, while retaining over-parameterization benefits during fine-tuning.

\subsection{Over-parameterized low-rank matrix selection}

The MPO decomposition method is recognized for its computational tractability and representational flexibility. However, its application for over-parameterizing all low-rank parameter matrices remains computationally expensive. To concentrate the benefits of over-parameterization on the critical parameters, the approach selectively applies MPO decomposition only to the important low-rank parameter matrix. Consequently, a dual strategy is proposed: a predefined selection method, which preidentifies a significant low-rank parameter matrix prior to fine-tuning, and a runtime selection method, which continuously identifies and selects critical low-rank parameter matrices throughout the fine-tuning process.

\paragraph{Predefined selection strategy.}
The proposed predefined selection strategy involves pre-computing importance scores for all candidate low-rank parameter matrices prior to the fine-tuning phase. Subsequently, this strategy employs the MPO formalism to over-parameterize the top-$N$ low-rank parameter matrices exclusively. Consequently, the architecture of the resulting over-parameterized LoRA model remains fixed throughout the subsequent fine-tuning process. Inspired by network pruning methods \citep{molchanov2016pruning,voita2019analyzing}, we quantify importance scores for individual low-rank parameter matrices by measuring the resulting perturbation in training loss $\mathcal{L}_\mathbf{W}$ after surgical removal of each low-rank parameter matrix $\mathbf{W}$. This metric is theoretically grounded in the principle that parameters exerting significant influence on predictive accuracy will inherently manifest elevated loss differentials when excised, as such a low-rank parameter matrix fundamentally supports the correct assignment of labels \citep{voita2019analyzing}. Thus, the importance score $I_\mathbf{W}$ of a low-rank parameter matrix $\mathbf{W}$ can be calculated as:
where $\mathcal{L}_\mathbf{W=0}$ represents the loss value after zeroing $\mathbf{W}$. To compute the loss, fine-tuning must commence from the same pre-trained parameters as our baseline, prior to integrating the low-rank parameter matrix. Crucially, low-rank parameter matrices originating from distinct modules inherently vary in size and functionality, rendering direct performance comparisons invalid. To address this methodological challenge, we first classify all low-rank parameter matrices into module-specific categories, where each group corresponds to a single modular component spanning $L$ layers. Within each such group, the top $N$ performing low-rank parameter matrices are subsequently isolated for over-parameterization.
\begin{equation}
    I_\mathbf{W} = |\mathcal{L_\mathbf{W}}-\mathcal{L}_\mathbf{W=0}|, \label{diff}
\end{equation}

\paragraph{Runtime selection strategy.}
We propose a runtime selection strategy that continuously computes importance scores to identify a significant low-rank parameter matrix for instantaneous over-parameterization during fine-tuning. This approach dynamically assesses the importance of change with respect to the optimization of the whole model. The approximation of the importance score can be obtained by performing the first-order Taylor expansion on \Cref{diff}:
\begin{equation}
    I_{\mathbf{W}} =\left|\mathcal{L}_{\mathbf{W}}-\left(\mathcal{L}_{\mathbf{W}}-\frac{\partial \mathcal{L}}{\partial \mathbf{W}}(\mathbf{W}-\mathbf{0})+R_{\mathbf{W}=\mathbf{0}}\right)\right|  
    \approx\left|\frac{\partial \mathcal{L}}{\partial \mathbf{W}} \mathbf{W}\right|. \label{score}
\end{equation}
The omission of part $R_\mathbf{W=0}$ allows the computation of the important score through the absolute gradients of the parameter matrix. Throughout the fine-tuning process, accumulation occurs for the absolute gradients across all low-rank parameter matrices. Dynamically computed using the \Cref{score}, importance scores trigger over-parameterization of top-$N$ low-rank parameter matrices in categorized groups at $t$-steps. The iterative cycle of this process persists until the selection of $N$ low-rank parameter matrices is achieved per group. Moreover, we also present a detailed algorithm for our selection strategy (see \Cref{alg:matrices_selection} in \Cref{appendix_algorithm}). As shown in \Cref{tab:nlu_results} and \Cref{tab:LLMs_results}, this strategy substantially improves fine-tuning performance. 

%% file: section/sec-experiment.tex
\section{Experiments}
\label{experiments}

In this section, we assess the performance of LoRA-Over on various benchmark datasets. Initially, we employ comprehensive experiments on the GLUE benchmark \citep{wang2018glue} with the T5-Base \citep{raffel2020exploring} model. Subsequently, we evaluate dialogue, arithmetic reasoning, and coding abilities using the Llama 2-7B \citep{touvron2023llama} model and the Llama 3.1-8B \citep{grattafiori2024llama} model. Finally, we give a detailed analysis. The implementation details can be found in the \Cref{appendix: experiment detail}. All experiments used single-epoch training with three seeds, reporting
mean values with standard deviations.

\subsection{Experiments on natural language understanding}
\input{section/table/nlu_results}
\paragraph{Models and datasets.}
To present a comprehensive overview of the performance of our proposed LoRA-Over, we fine-tune the T5-Base model on a subset of GLUE datasets, including MNLI, SST-2, CoLA, QNLI, and MRPC. Performance is evaluated on the corresponding validation sets using accuracy as the metric.

\paragraph{Results.}
As shown in \Cref{tab:nlu_results}, LoRA-Over-MPO$_R$ achieves superior performance on most datasets. It achieves the highest accuracy on SST-2 (94.50), CoLA (81.21), QNLI (93.26), and MRPC (85.38). While slightly underperforming LoRA+ on
MNLI by just 0.02 percentage points. Moreover, the average score of LoRA-Over-MPO$_R$ (88.03) surpasses all other methods and exceeds vanilla LoRA with a margin of 5.95, demonstrating outstanding adaptability and generalization. These results show the effectiveness of our approach. Notably, for small datasets, CoLA and MRPC, our method shows highly strong performance, highlighting that it utilizes small training data effectively.

\subsection{Experiments on natural language generation}
\input{section/table/LLMs_results}
\paragraph{Models and datasets.}
We evaluate the performance of LoRA-Over on Llama 2-7B and Llama 3.1-8B. For dialogue generation, we train our model on a 52k subset of the WizardLM dataset \citep{xu2024wizardlm} and evaluate it using the MT-Bench dataset. The quality of the model responses is assessed using GPT-4, with the first-turn score reported as the evaluation metric. For mathematical reasoning, we train our model on a 100k subset of MetaMathQA \citep{yu2023metamath}. The model's performance is evaluated on the GSM8K test set, with accuracy reported as the metric. For code generation, we train our model on a 100k subset of the CodeFeedback dataset \citep{zheng2024opencodeinterpreter} and evaluate it on the HumanEval dataset, and the model performance is quantified via the PASS@1 metric.

\paragraph{Results.}
The results presented in \Cref{tab:LLMs_results} reveal a consistent performance advantage of LoRA-Over over other baseline methods in most tasks. For Llama 2-7B, LoRA-Over-MPO$_R$ demonstrates exceptional performance on both the GSM8K and HumanEval datasets. Although slightly underperforming DoRA on MT-Bench by just 0.05 points, LoRA-Over-MPO$_R$'s average score (26.70) surpasses all baselines. Specifically, LoRA-Over-MPO$_R$ achieves the highest score on GSM8K (54.21) and HumanEval (19.96), substantially
surpassing vanilla LoRA, indicating superior performance in mathematical reasoning and code generation. For Llama 3.1-8B, LoRA-Over-MPO$_R$ exhibits superior performance compared to the baseline methods. It achieves the highest score on GSM8K (73.95), achieving performance better than full finetuning (73.69). On MT-Bench, LoRA-Over-MPO$_R$ scores 6.26, placing second to LoRA+. While on HumanEval, it ranks second with 44.84, just behind rsLoRA. The average performance of LoRA-Over-MPO$_R$ is 41.68, surpassing LoRA+'s 40.72 by 0.96 points. Remarkably, LoRA-Over-MPO$_R$ demonstrates robust performance across various rank allocation settings. For example, in Llama-2-7B and Llama-3.1-8B model, our method achieves superior performance at both $r=32$ and $r=128$. The results indicate that SVD typically demonstrates inferior performance compared to MPO among the two matrix decomposition methods. We observe that the runtime strategy generally exhibits superior performance over the predefined strategy at identical parameter scales. Empirical evidence suggests that LoRA-Over-MPO$_R$ substantially enhances training stability and consistently yields robust performance across diverse natural language generation tasks.

\subsection{Further analysis}

We then conduct a comprehensive analysis to rigorously evaluate the proposed approach. Particularly, the computational cost analysis can be found in \Cref{appendix:cost analysis}.
\paragraph{Parameter efficiency analysis.}
We compare LoRA-Over-MPO$_R$ with vanilla LoRA under an iso-parameter setting to ensure a rigorous evaluation. Results are shown in \Cref{tab:iso}. By scaling the rank of LoRA to match our parameter count, we observe that LoRA-Over-MPO$_R$ achieves superior performance. Even when vanilla LoRA is scaled to have the same number of trainable parameters, our method consistently outperforms it, demonstrating superior parameter efficiency.\input{section/table/iso-prarameter}

\begin{figure*}[h]
\centering
  \includegraphics[width=1\linewidth]{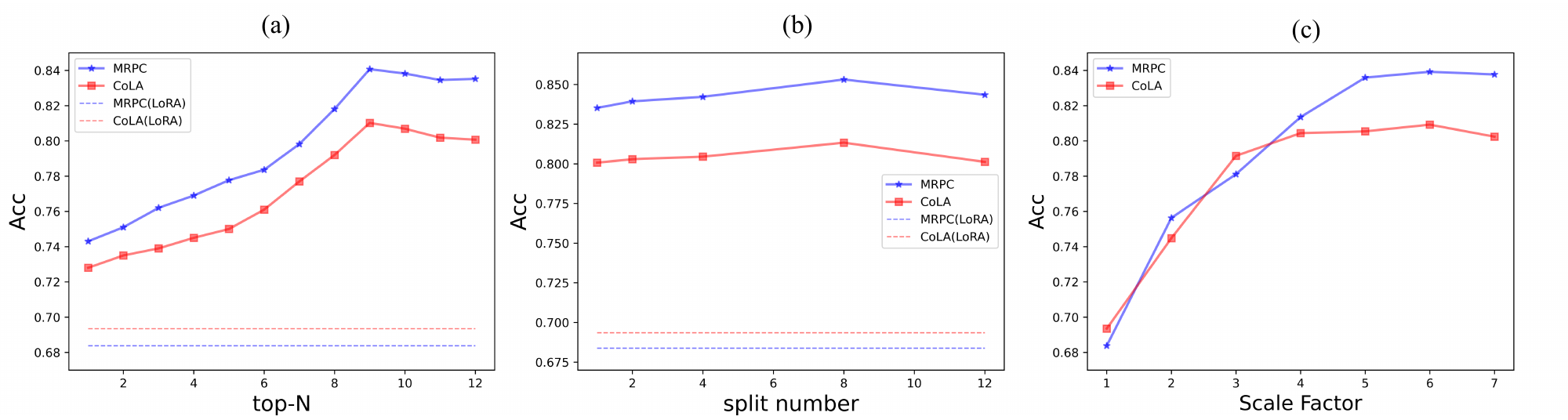}
  \caption{Performance comparison with varying hyper-parameters on MRPC and CoLA using T5-Base. (\textbf{a})Hyper-parameter: top-$N$. (\textbf{b})Hyper-parameter: split number. (\textbf{c})Performance comparison under different parameter expansion scales on MRPC and CoLA tasks.
  }
  \label{fig:analysis}
\end{figure*}
\paragraph{Hyper-parameters tuning.}
The performance of our method is primarily influenced by two hyper-parameters: the total count of selected parameter matrices $N$ and split number, making their tuning crucial. particularly, a higher value of $N$ corresponds to a larger number of parameter matrices being selected and over-parameterized, while a smaller split number indicates that a larger subset of matrices is over-parameterized simultaneously in a single operation. To investigate the effects of these hyperparameters, we perform an empirical analysis on the CoLA and MRPC datasets using the T5-base model. As illustrated in \Cref{fig:analysis}(a), for LoRA-Over-MPO$_P$, model performance improves steadily with increasing values of $N$ before eventually saturating. This trend suggests that an insufficient degree of over-parameterization fails to adequately capture the full representation capacity of the low-rank matrix. As illustrated in \Cref{fig:analysis}(b), for LoRA-Over-MPO$_R$, we observe that performance scales positively with split number but diminishes when split number exceeds an optimal threshold.

\paragraph{Parameter increasing rate analysis.}
To enhance the over-parameterization of low-rank matrices, our method deliberately increases the number of trainable parameters during fine-tuning. Given that the proposed method provides a general and flexible framework for scaling trainable parameters, we systematically evaluate its performance across multiple parameter counts. Based on LoRA's model of T5-Base, we systematically increase the trainable parameters via over-parameterization (from $1 \times$ to $7 \times$) and assess the performance of LoRA-Over-MPO$_P$ on the MRPC and CoLA tasks. \Cref{fig:analysis}(c) demonstrates a consistent, monotonic improvement in model performance as the parameter scale increases. Empirical results indicate that performance improvement asymptotically approaches a plateau at the $7 \times$ parameter scale. A potential interpretation is that this scale exhausts the primary benefits of over-parameterization for this specific architecture and task set.

%% file: section/table/nlu_results.tex
\begin{table*}[htb!]
\small
    \centering
    \caption{Performance comparison using T5-Base on the GLUE benchmark (in percent). \textbf{Bold} represents the best performance, \underline{underline} indicates the second-best performance.}
        \resizebox{.8\columnwidth}{!}{
            \begin{tabular}{l|ccccc|c}
            \toprule
                \textbf{Method} & \textbf{MNLI} & \textbf{SST2} & \textbf{CoLA} & \textbf{QNLI} & \textbf{MRPC} & \textbf{Avg} \\
                \midrule
                \rowcolor{gray!10} \textit{Train Size} & \textit{393k} & \textit{67k} & \textit{8.5k} & \textit{105k} & \textit{3.7k} & \\
                \midrule
                Full & \(86.33_{\pm 0.00}\) & \(94.75_{\pm 0.21}\) & \(80.70_{\pm 0.24}\) & \(93.19_{\pm 0.22}\) & \(84.56_{\pm 0.73}\) & \(87.91\) \\
                LoRA & \(85.30_{\pm 0.04}\) & \(94.04_{\pm 0.11}\)  & \(69.35_{\pm 0.05}\)  &  \(92.96_{\pm 0.09}\)  & \(68.38_{\pm 0.01}\)  &  \(82.08\)\\
                \midrule
                \rowcolor{gray!20}\multicolumn{7}{c}{\textit{LoRA Variants with Original Structure}}    \\
                \midrule
                PiSSA & \({85.75}_{\pm 0.07}\)& \(94.07_{\pm 0.06}\) & \(74.27_{\pm 0.39}\)  & \({93.15}_{\pm 0.14}\)   & \({76.31}_{\pm 0.51}\)& \(84.71\) \\
                rsLoRA & \(85.73_{\pm 0.10}\) & \(94.19_{\pm 0.23}\)  & \(72.32_{\pm 1.12}\) & \(93.12_{\pm 0.09}\)   & \(52.86_{\pm 2.27}\)  & \(79.64\) \\
                LoRA+ & \(\mathbf{85.81}_{\pm 0.09}\) & \(93.85_{\pm 0.24}\)  & \({77.53}_{\pm 0.20}\) & \(93.14_{\pm 0.03}\)   & \(74.43_{\pm 1.39}\)  & \({84.95}\)\\
                LoRA-GA & 85.70$_{\pm0.09}$ & 94.11$_{\pm0.18}$ & 80.57$_{\pm0.20}$ & 93.18$_{\pm0.06}$ & \underline{85.29$_{\pm0.24}$} & \underline{87.77} \\
                \midrule 
                \rowcolor{gray!20}\multicolumn{7}{c}{\textit{LoRA Variants with Modified Structure}}    \\
                \midrule
                DoRA & \(85.67_{\pm 0.09}\) & \(94.04_{\pm 0.53}\)  & \(72.04_{\pm 0.94}\) & \(93.04_{\pm 0.06}\)  & \(68.08_{\pm 0.51}\) &  \(82.57\) \\
                AdaLoRA & \(85.45_{\pm 0.11}\)  & \(93.69_{\pm 0.20}\) & \(69.16_{\pm 0.24}\) & \(91.66_{\pm 0.05}\) & \(68.14_{\pm 0.28}\)  & \(81.62\) \\
                \midrule
                LoRA-Over-SVD & 85.33$_{\pm 0.06}$ & 94.13$_{\pm 0.29}$ & 74.52$_{\pm 0.39}$ & 93.05$_{\pm 0.11}$ & 75.76$_{\pm 0.23}$ & 84.56 \\
                LoRA-Over-MPO & 85.42$_{\pm 0.09}$ & 94.35$_{\pm 0.32}$ & 80.79$_{\pm 0.39}$ & 93.14$_{\pm 0.39}$ & 82.84$_{\pm 0.53}$ & 87.31 \\
                LoRA-Over-MPO$_P$ & 85.59$_{\pm 0.08}$ & \underline{94.41$_{\pm 0.14}$} & \underline{81.02$_{\pm 0.16}$} & \underline{93.19$_{\pm 0.05}$} & 84.07$_{\pm 0.69}$ & 87.66 \\
                LoRA-Over-MPO$_R$ & \underline{85.79$_{\pm 0.12}$} & \(\mathbf{94.50}_{\pm 0.58}\) & \(\mathbf{81.21}_{\pm 0.41}\) & \(\mathbf{93.26}_{\pm 0.08}\) & \(\mathbf{85.38}_{\pm 1.22}\) & \(\mathbf{88.03}\) \\
                \bottomrule
                \end{tabular}
        }
    \label{tab:nlu_results}
\end{table*}

%% file: section/table/LLMs_results.tex
\begin{table*}[ht!]
\small
    \centering
    \caption{ Performance comparison using Llama 2-7B and Llama 3.1-8B on MT-Bench, GSM8K, and HumanEval (in percent). \textbf{Bold} and \underline{underline} indicate the highest and second-highest scores, respectively.}
        \resizebox{.99\columnwidth}{!}{
            \begin{tabular}{c|c|ccc|c}
                \bottomrule
                 \textbf{LLM} & \textbf{Method} & \textbf{MTBench} & \textbf{GSM8K} & \textbf{HumanEval} &
                 \textbf{Avg}\\
                \hline
                & Full & 5.30$_{\pm0.11}$ & 59.36$_{\pm0.85}$ & 35.31$_{\pm2.13}$ & 33.32 \\
                & LoRA & 5.61$_{\pm0.10}$ & 42.08$_{\pm0.04}$ & 14.76$_{\pm0.17}$ & 20.82 \\
                \cline{2-6}
                & PiSSA & 5.30$_{\pm0.02}$ & 44.54$_{\pm0.27}$ & 16.02$_{\pm0.78}$ & 21.95 \\
                & rsLoRA & 5.25$_{\pm0.03}$ & 45.62$_{\pm0.10}$ & 16.01$_{\pm0.79}$ & 22.29 \\
                & OLoRA & 5.30$_{\pm0.04}$ & 43.29$_{\pm0.83}$ & 17.22$_{\pm0.12}$ & 21.94 \\
                & LoRA+ & 5.71$_{\pm0.08}$ & 52.11$_{\pm0.62}$ & 18.17$_{\pm0.52}$ & 25.33 \\
                & LoRA-GA & \underline{5.95$_{\pm0.16}$} & \underline{53.60$_{\pm0.30}$} & \underline{19.81}$_{\pm1.46}$ & \underline{26.45}  \\
                & AdaLoRA & 5.57$_{\pm0.05}$ & 50.72$_{\pm1.39}$ & 17.80$_{\pm0.44}$ & 24.70 \\
                & DoRA & \(\mathbf{5.97}_{\pm0.02}\) & 53.07$_{\pm0.75}$ & 19.75$_{\pm0.41}$ & 26.26 \\
                \cline{2-6}
                & LoRA-Over-SVD & 5.23$_{\pm0.08}$ & 45.01$_{\pm0.54}$ & 15.22$_{\pm0.27}$ & 21.82
                \\
                & LoRA-Over-MPO & 5.63$_{\pm0.13}$ & 49.51$_{\pm0.82}$ & 17.32$_{\pm0.49}$ & 24.15
                \\
                & LoRA-Over-MPO$_P$ & 5.66$_{\pm0.09}$ & 50.13$_{\pm0.73}$ & 17.68$_{\pm0.31}$ & 24.49
                \\
                & LoRA-Over-MPO$_R$ & 5.92$_{\pm0.07}$ & \(\mathbf{54.21}_{\pm0.66}\) & \(\mathbf{19.96}_{\pm0.25}\) & \(\mathbf{26.70}\)
                \\
                & LoRA-Over-MPO$_R$(rank=32) & 5.81$_{\pm0.05}$ & 55.46$_{\pm0.58}$ & 20.83$_{\pm0.44}$ & 27.37 \\
                \multirow{-15}{*}{Llama 2-7B \citep{touvron2023llama}} & LoRA-Over-MPO$_R$(rank=128) & 5.94$_{\pm0.06}$ & 56.13$_{\pm0.87}$ & 23.65$_{\pm0.51}$ & 28.57 \\
                \hline
                & Full & 5.88$_{\pm0.23}$ & 73.69$_{\pm0.28}$ & 51.63$_{\pm1.27}$ & 43.73 \\
                & LoRA & 6.15$_{\pm0.02}$ & 67.78$_{\pm1.25}$ & 43.09$_{\pm0.35}$ & 39.01 \\
                \cline{2-6}
                & PiSSA & 6.08$_{\pm0.09}$ & 68.56$_{\pm1.03}$ & 44.10$_{\pm1.54}$ & 39.58\\
                & rsLoRA & 6.18$_{\pm0.09}$ & 68.36$_{\pm0.74}$ & \(\mathbf{45.78}_{\pm2.80}\) & 40.11 \\
                & OLoRA & 6.13$_{\pm0.04}$ & 68.54$_{\pm0.42}$ & 43.29$_{\pm2.44}$ & 39.32 \\
                & LoRA+ & \(\mathbf{6.35}_{\pm0.10}\) & 71.29$_{\pm0.93}$ & 44.51$_{\pm2.11}$ & \underline{40.72} \\
                & LoRA-GA & 5.99$_{\pm0.06}$ & 71.39$_{\pm0.90}$ & 43.29$_{\pm0.61}$ & 40.22  \\
                & AdaLoRA & 6.19$_{\pm0.16}$ & 70.63$_{\pm0.77}$ & 41.46$_{\pm3.66}$ & 39.43 \\
                & DoRA & 6.24$_{\pm0.12}$ & 69.17$_{\pm1.00}$ & 43.70$_{\pm1.54}$ & 39.70 \\
                \cline{2-6}
                & LoRA-Over-SVD & 6.01$_{\pm0.04}$ & 68.92$_{\pm0.34}$ & 42.60$_{\pm0.37}$ & 39.18
                \\
                & LoRA-Over-MPO & 6.12$_{\pm0.07}$ & 72.14$_{\pm0.16}$  & 43.15$_{\pm0.24}$ & 40.47
                \\
                & LoRA-Over-MPO$_P$ & 6.16$_{\pm0.11}$ & \underline{72.73$_{\pm0.10}$}  & 43.34$_{\pm0.26}$ & 40.74
                \\
                & LoRA-Over-MPO$_R$ & \underline{6.26$_{\pm0.08}$} & \(\mathbf{73.95}_{\pm0.42}\) & \underline{44.84$_{\pm1.54}$} & \(\mathbf{41.68}\) \\
                & LoRA-Over-MPO$_R$(rank=32) & 6.23$_{\pm0.06}$ & 74.98$_{\pm0.25}$& 45.44$_{\pm1.09}$ & 42.22 \\
                \multirow{-15}{*}{Llama 3.1-8B \citep{grattafiori2024llama}} & LoRA-Over-MPO$_R$(rank=128) & 6.31$_{\pm0.12}$ & 75.29$_{\pm0.56}$ & 46.15$_{\pm1.32}$ & 42.58 \\
                \hline
                \end{tabular}
        }
    \label{tab:LLMs_results}
\end{table*}

%% file: section/table/iso-prarameter.tex
\begin{table*}[ht!]
\small
\caption{ Comparison with vanilla LoRA under equivalent parameter budgets.}
    \centering
        \resizebox{.85\columnwidth}{!}{
            \begin{tabular}{c|c|ccc|c}
                \bottomrule
                 \textbf{LLM} & \textbf{Method} & \textbf{MTBench} & \textbf{GSM8K} & \textbf{HumanEval} &
                 \textbf{Avg}\\
                \hline
                & LoRA(rank=34) & 5.89$_{\pm0.06}$ & 48.07$_{\pm0.29}$ & 17.32$_{\pm0.31}$ & 23.76 \\
                \cline{2-6}
                \multirow{-2}{*}{Llama 2-7B \citep{touvron2023llama}} & LoRA-Over-MPO$_R$ & \(\mathbf{5.92}_{\pm0.07}\) & \(\mathbf{54.21}_{\pm0.66}\) & \(\mathbf{19.96}_{\pm0.25}\) & \(\mathbf{26.70}\)
                \\
                \hline
                & LoRA(rank=42) & 6.23$_{\pm0.06}$ & 72.24$_{\pm0.32}$ & 43.66$_{\pm0.42}$ & 40.71 \\
                \cline{2-6}
                \multirow{-2}{*}{Llama 3.1-8B \citep{grattafiori2024llama}} & LoRA-Over-MPO$_R$ & \(\mathbf{6.26}_{\pm0.08}\) & \(\mathbf{73.95}_{\pm0.42}\) & \(\mathbf{44.84}_{\pm1.54}\) & \(\mathbf{41.68}\) \\
                
                \hline
                \end{tabular}
        }
    \label{tab:iso}
\end{table*}

%% file: section/sec-conclusion.tex
\section{Conclusion and limitation}
\paragraph{Conclusion.}
In this paper, we introduce LoRA-Over, a novel framework that enhances low-rank adaptation through structured over-parameterization. By leveraging MPO decomposition to augment matrix capacity and employing adaptive selection strategies (predefined and runtime), LoRA-Over effectively prioritizes parameter expansion for task-critical layers. Our extensive experiments demonstrate that this approach significantly boosts the performance of LoRA, effectively narrowing the gap between LoRA and full fine-tuning while maintaining architectural efficiency.
\paragraph{Limitation.}While LoRA-Over shows promise, it has limitations. First, the evaluation focuses on a limited range of NLP tasks, leaving its generalizability to broader domains unverified. Second, we have not tested the framework on ultra-large-scale models due to resource constraints, which may exhibit different scaling behaviors.

%% file: section/sec-appendix.tex
\clearpage
\newpage
\appendix

\section{Details of tensors}
\label{appendix: detail of tensor}
\subsection{Tensor and matrix product operators}
As introduced in \citep{gao2020compressing}, a tensor is precisely characterized as follows:
\paragraph{Tensor.}
Let $D_1, D_2,\cdots,D_P \in N$ denote index upper bounds. A tensor $\mathcal{T} \in \mathbb{R}^{D_1, D_2,\cdots,D_P}$ of order $P$ is a $P$-way array where elements $\mathcal{T}[d_1,d_2,\cdots,d_P]$ are indexed by $d_p \in \{1,2,\cdots,D_P\}$ for $1 \leq p \leq P$.

\paragraph{Matrix product operator.}
The concept of bond dimension $d_k$ is defined as \Cref{d_k}. We can observe that is will be large in the middle and small on both sides by \Cref{d_k}. The MPO representation decomposes $M$ into a product of $n$ local tensors:
\begin{equation}
    M_{i_1\cdots i_n,j_1\cdots j_n} = \mathcal{T}^{(1)}[i_1,j_1]\cdots \mathcal{T}^{(n)}[i_n,j_n],
\end{equation}
where $\mathcal{T}^{(p)}[i_p,j_p]$ is a $D_{p-1} \times D_p$ matrix with $D_p$ the virtual basis dimension on the bond linking $\mathcal{T}^{(p)}$ and $\mathcal{T}^{(p+1)}$ with $D_0=D_n=1$.
\label{appendix:tensor and mpo}

\subsection{Theorem}
\label{appendix:theorem}
Suppose that the tensor \textbf{W}$^{(k)}$ of matrix $W$ that is satisfy:
\begin{equation}
    \mathbf{W} = \mathbf{W}^{(k)}+\mathbf{E}^{(k)},D(\boldsymbol{W}^{(k)})=d_k,
\end{equation}
where $\lVert \mathbf{E}^{(k)} \rVert^2_F=\epsilon^2_k,k=1,\cdots,d-1$.
Then MPO($\mathbf{W}$) with the $k$-th bond dimension $d_k$ upper bound of truncation error satisfy:
\begin{equation}
    \lVert \mathbf{W}-\mathrm{MPO}(\mathbf{W}) \rVert_F \leq \sqrt{\sum^{d-1}_{k=1}\epsilon^2_k}.
\end{equation}
$Proof.$ The proof is by induction. For $n=2$ the statement follows from the properties of the SVD. Consider an arbitrary $n>2$. Then the first unfolding $\mathbf{W}^{(1)}$  is decomposed as:
\begin{equation}
    \mathbf{W}^{(1)} = \mathbf{U}_1\lambda_1\mathbf{V}_1+\mathbf{E}^{(1)}=\mathbf{U}_1\mathbf{B}^{(1)}+\mathbf{E}^{(1)},
\end{equation}
where $\mathbf{U}_1$ is of size $r_1 \times i_1 \times j_1$ and $\lVert \mathbf{E}^{(1)} \rVert^2_F=\epsilon^2_1$. The matrix $\mathbf{B}_1$ is naturally associated with a $(n-1)$-dimensional tensor $\mathcal{B}^{(1)}$ with elements $\mathcal{B}^{(1)}(\alpha,i_2,j_2,\cdots,i_n,j_n)$, which will be decomposed further. This means that $\mathbf{B}_1$ will be approximated by some other matrix $\hat{\mathbf{B}_1}$. From the properties of
the SVD it follows that $\mathbf{U}^{\mathrm{T}}_1\mathbf{E}^{(1)}=0$, and thus:
\begin{align}
&\lVert\mathbf{W}-\mathcal{B}^{(1)}\rVert_{F}^{2} \notag\\
&= \lVert\mathbf{W}_1-\mathbf{U}_1\hat{\mathbf{B}_1}\rVert_{F}^{2}\notag\\
&= \lVert\mathbf{W}_1-\mathbf{U}_1(\hat{\mathbf{B}_1}+\mathbf{B}_1-\mathbf{B}_1)\rVert_{F}^{2}\notag\\
&= \lVert\mathbf{W}_1-\mathbf{U}_1\mathbf{B}_1\rVert_{F}^{2}+\lVert\mathbf{U}_1(\hat{\mathbf{B}_1}-\mathbf{B}_1)\rVert_{F}^{2},
\end{align}
and since $\mathbf{U}_1$ has orthonormal columns,
\begin{equation}
    \lVert\mathbf{W}-\mathcal{B}^{(1)}\rVert_{F}^{2} \leq \epsilon^2_1 + \lVert\mathbf{B}_1-\hat{\mathbf{B}_1}\rVert_{F}^{2},\label{appendix.s7}
\end{equation}
and thus it is not difficult to see from the orthonormality of columns of $\mathbf{U}_1$ that the distance of the $k$-th unfolding $(k=2,\cdots,d_k-1)$ of the $(d-1)$-dimensional tensor $\mathcal{B}^{(1)}$
 to the $d_k$-th rank matrix cannot be larger than $\epsilon_k$. Proceeding by induction, we have
 \begin{equation}
     \lVert\mathbf{B}_1-\hat{\mathbf{B}_1}\rVert_{F}^{2} \leq \sum ^{d-1}_{k=2} \epsilon^2_K,
 \end{equation}
combine with \Cref{appendix.s7}, this completes the proof.

\section{Algorithms}
\label{appendix_algorithm}
The MPO pseudocode is shown in \Cref{alg:mpo}. 
\input{section/algorithm/mpo}
The pseudocode for the selection of over-parameterized matrices is shown in \Cref{alg:matrices_selection}
\input{section/algorithm/matrices_selection}

\section{Additional experiment details}
\label{appendix: experiment detail}
In this paper, we propose the MPO decomposition as a method to increase the parameters of the model. Using \Cref{mpo_define}, an MPO can be specified as: 
\begin{equation}
    \mathcal{T}^{j_1,j_2,\cdots,j_n}_{i_1,i_2,\cdots,i_n}(D).
\end{equation}
We pre-compute the significance scores of all parameter matrices before fine-tuning and subsequently over-parameterize the top $N$ ones using the MPO technique. The significance score can be calculated by \Cref{diff} and \Cref{score}.

\subsection{Baseline methods}
\label{appendix:baseline}
We compare LoRA-Over with several baselines to demonstrate its effectiveness. \textit{Full fine-tuning} is the most prevalent adaptation approach, which updates all model parameters but requires substantial computational resources. \textit{LoRA} \citep{hu2022lora} constitutes a parameter-efficient fine-tuning method that introduces a low-rank parameter matrix product $BA$ into linear layers, where $A$ is initialized by Kaiming initialization and $B$ is initialized to zero.  \textit{rsLoRA} \citep{kalajdzievski2023rank} incorporates a novel scaling factor to enhance the stability of the LoRA scale. \textit{PiSSA} \citep{meng2024pissa} employs SVD \citep{henry19928} in the weight matrix $W$ at initialization, and uses larger singular values for better performance. \textit{LoRA+} \citep{hayou2024lora+} applies different learning rates for the low-rank matrix $A$ and $B$ in LoRA. \textit{OLoRA} \citep{buyukakyuz2024olora} incorporates orthonormal initialization for the adaptation matrices. \textit{DoRA} \citep{liu2024dora} enhances expressiveness by adding learnable magnitudes. \textit{AdaLoRA} \citep{zhang2023adalora} dynamically prunes nonessential weights via SVD,  reallocating rank to diminish GPU memory. Furthermore, we conduct a comparative analysis between our method and SVD, a classical matrix factorization technique, which is viable for over-parameterizing our model. Concretely, this methodological substitution replaces MPO with SVD within our method, implementing comprehensive over-parameterization across all the low-rank parameter matrices during fine-tuning. 

\subsection{Implementation details}
\label{implementation_details}
To ensure a fair comparison, we maintain the experimental setup of GoRA \citep{he2025gora} and adopt baseline performances reported by them. By default, we fine-tune the converged models using the AdamW optimizer \citep{loshchilov2017decoupled} with $\beta_1=0.9$, $\beta_2=0.999$, $\epsilon=1e-8$. We implement a cosine learning rate schedule with a warmup ratio of 0, setting the rank $r=8$ and $\alpha = 16$. For natural language understanding tasks, we fine-tune T5-base \citep{raffel2020exploring} with a sequence length of 128, a training batch size of 32, and a weight decay of 0. For natural language generation tasks, we fine-tune Llama 2-7B \citep{touvron2023llama} with a sequence length of 1024, a training batch size of 32, a weight decay of 0, and fine-tune Llama 3.1-8B \citep{grattafiori2024llama} with a sequence length of 512, a training batch size of 64, a weight decay of $5e-4$. All generation is performed with $top\_p=0.95$ and temperature $T=0.8$. For T5-Base and Llama 2-7B, the LoRA target is all linear modules except the embedding layer, layer norm, and language model head. For Llama 3.1-8B, we train the linear components of the attention modules. In our work, we adopt two types of GPUs: NVIDIA H100 GPU and NVIDIA H200 GPU. For the Natural Language Understanding tasks, all computations are performed on a single H200 GPU. For the Natural Language Generation tasks, computations for the Llama 2-7B model are performed on a single H100 GPU, while the Llama 3.1-8B model are processed on a  single H200 GPU. The hyperparameters of LoRA-Over using the T5-Base model and the MPO structure of the T5-Base model are presented in the \Cref{tab:hyper_glue} and the \Cref{tab:mpo_structure_T5-base}, respectively.
The hyperparameters of LoRA-Over using Llama 2-7B and Llama 3.1-8B models are presented in the \Cref{tab:hyper_LLMs}. The MPO structure of Llama 2-7B and Llama 3.1-8B model is presented in the \Cref{tab:mpo_structure_llama}.

\input{section/table/hyper_glue}
\input{section/table/mpo_structure_T5_Base}

\input{section/table/hyper_LLMs}
\input{section/table/mpo_structure_llama}

\section{Discussion}
\label{appendix:discussion}
We position our MPO-based approach within the broader landscape of low-rank approximation techniques, systematically comparing it with SVD \citep{henry19928}, CPD \citep{hitchcock1927expression}, and Tucker decomposition \citep{tucker1966some}, as shown in \Cref{tab-comparison}. To facilitate a rigorous comparison, we characterize the inference time complexities of these methods using a unified set of parameters. Conceptually, these techniques can be categorized into two primary families based on their structural properties: the Tucker family and the MPO family. In this hierarchy, CPD serves as a constrained special case of Tucker decomposition, where the core tensor is strictly diagonal, while SVD represents the degenerate case of MPO when the tensor order is reduced to $n = 2$. A critical distinction lies in their scalability, Tucker decomposition suffers from exponential complexity growth with respect to the tensor order $n$ due to its dense core, while MPO maintains a linear scaling behaviour. Consequently, as demonstrated in \Cref{tab-comparison}, MPO achieves significantly lower inference complexity than Tucker decomposition for higher-order tensors $(n > 3)$, making it a more efficient choice for complex multi-dimensional data.

\begin{table}[htb!]
\caption{Comparison of inference time complexities for various low-rank approximation methods. The notation is defined as follows: $n$ is the tensor count, $m=\max(\{i_k\}_{k=1}^n)$ is the largest input dimension, and $d=\max(\{d_k\}_{k=0}^n)$ is the maximum rank.}
\centering
\resizebox{.65\columnwidth}{!}{%
\begin{tabular}{clll}
\toprule[1pt]
Method Family & Specific Case & Condition & Inference Complexity \\ \midrule[0.7pt]
\multirow{2}{*}{Tucker}    & CPD & $d=1$ &   $\mathcal{O}(nmd^2)$             \\
                           & General Tucker  & $d>1$  &    $\mathcal{O}(nmd+d^n)$             \\ \midrule[0.7pt]
\multirow{2}{*}{MPO}       & SVD   &  $n=2$ & $\mathcal{O}(2md^3)$           \\
                           & General MPO   & $n>2$  &    $\mathcal{O}(nmd^3)$             \\
\bottomrule[1pt]
\end{tabular}%
}
\label{tab-comparison}
\end{table}
Our methodology is grounded in the theoretical foundation established by OPF \citep{gao2023small}, which posits that over-parameterization significantly enhances the optimization process during fine-tuning. However, a critical distinction lies in the migration of the application domain and the expansion of the theoretical boundaries. While OPF primarily explores the benefits of over-parameterization within general fine-tuning regimes, our work focuses on validating and realizing the efficacy of this theory under the specific constraints of low-rank approximation. Specifically, unlike the original study, which targets performance improvements in full-parameter or high-degree-of-freedom settings, we narrow our focus to the strictly constrained hypothesis space characteristic of LoRA and its variants. Our empirical results demonstrate that, even within a low-rank framework where the parameter space is drastically compressed, the over-parameterization mechanism remains robust, thereby enhancing feature representation by smoothing the optimization landscape. This constitutes not only a substantial extension of the OPF theory into the realm of PEFT but also a seamless integration with existing LoRA-based methods, thereby breaking through the performance bottlenecks of lightweight fine-tuning while maintaining strict control over the trainable parameter budget.
\section{Computational cost analysis}
\label{appendix:cost analysis}

\input{section/table/memory}
\input{section/table/parameter_count}
To comprehensively validate the efficiency of LoRA-Over-MPO$_R$, we evaluate its computational characteristics on a single NVIDIA H200 GPU, focusing on training latency and peak GPU memory footprint. First, regarding inference efficiency, our method introduces zero inference overhead. Since LoRA-Over-MPO$_R$ allows for the merging of learned weights back into the backbone model—similar to vanilla LoRA—it creates no additional architectural burden during deployment. Thus, despite the structural changes during training, the inference latency remains identical to the baseline. Second, in terms of memory consumption, \Cref{tab:memory} demonstrates that LoRA-Over-MPO$_R$ is significantly more memory-efficient. This advantage stems from our MPO decomposition strategy, which breaks down large weight matrices into a series of compact core tensors, thereby drastically reducing the activation and gradient memory footprint. Finally, regarding the training computation, while the tensor contraction operations induce a increase in training latency compared to LoRA, this is a justified trade-off. The substantial gains in fine-tuning performance and memory savings offset the increase in training time. To further illustrate the scalability of our approach, we benchmark the trainable parameter counts across diverse models and tasks, with detailed statistics provided in \Cref{tab:parameter_count}.

%% file: section/algorithm/mpo.tex
\floatname{algorithm}{Algorithm}
\begin{algorithm}[H]
\small
    \caption{MPO decomposition for a matrix.}   
    \begin{algorithmic}[1] 
        \Require matrix $\mathbf{M}$, the number of local tensors $m$.
        \Ensure : MPO tensor list $\{\mathcal{T}_{(s)}\}^m_{s=1}$.
        \For{$s=1 \to m$}
            \State $\mathbf{M}[I,J] \to \mathbf{M}[d_{s-1} \times i_s \times j_s,-1]$            
            \State $\mathbf{U}\lambda\mathbf{V}^{\mathrm{T}}=\mathrm{SVD}(\mathbf{M})$
            \State $\mathbf{U}[d_{s-1} \times i_s \times j_s,d_s] \to \mathcal{U}[d_{s-1},i_s,j_s,d_s]$
            \State $\mathcal{T}^{(s)} :=\mathcal{U}$
            \State $\mathbf{M}:=\lambda\mathbf{V}^{\mathrm{T}}$
        \EndFor \\
        $\mathcal{T}^{(s)}:=\mathbf{M}$ \\
        Normalization \\
        \Return{$\{\mathcal{T}_{(k)}\}^n_{k=1}$}
    \end{algorithmic}
\label{alg:mpo}
\end{algorithm}

%% file: section/algorithm/matrices_selection.tex
\floatname{algorithm}{Algorithm}
\begin{algorithm}[H]
\small
    \caption{Fine-tuning a model with our method.}   
    \begin{algorithmic}[1] 
        \Require Low-rank parameter matrices set of a model $\{\mathbf{W}\}$. \\
        Divide $\{\mathbf{W}\}$ into several groups by module.
        \If{is Predefined Strategy}
            \State LoRA fine-tuning the model until converged.
            \State Compute $I_{\mathbf{W}}$ for $\{\mathbf{W}\}$ using \Cref{diff}.
            \State Sort $\{\mathbf{W}\}$ in each group according to $I_{\mathbf{W}}$.
            \State Perform MPO on the top-$N$ matrices.
            \State Train the other PLM until converged.
        \Else
            \State Define $S=\{\}$
            \While{Len($S$)$<N$}
                \State Train the model for $t$ steps.
                \State Compute $I_{\mathbf{W}}$ for $\{\mathbf{W}\}$ using \Cref{score}.
                \State Sort $\{\mathbf{W}\}$ in each group according to $I_{\mathbf{W}}$.
                \State Add top-$n$ matrices into $S$, and perform MPO.
            \EndWhile
            \State Continually train the model until converged.
        \EndIf
    \end{algorithmic}
\label{alg:matrices_selection}
\end{algorithm}

%% file: section/table/hyper_glue.tex
\begin{table}[htb!]
\caption{Hyperparameter setup of LoRA-Over for GLUE benchmark (T5-Base). "LR" denotes the learning rate. "Null" denotes the parameter is useless.}
    \centering
        \resizebox{.8\columnwidth}{!}{
            \begin{tabular}{c|c|c|c|c|c|c}
                \hline
                 Method & Dataset & LR & MPO\_LR & split number & top-$N$ & eval step \\
                \hline
                & MNLI & 4.9e-4 & Null & Null & Null & 100 \\
                & QNLI & 4.9e-4 & Null & Null & Null & 80 \\
                & CoLA & 4.9e-4 & Null & Null & Null & 5 \\
                & SST-2 & 4.9e-4 & Null & Null & Null & 50 \\
                \multirow{-5}{*}{LoRA-Over-SVD} & MRPC & 4.9e-4 & Null & Null & Null & 5 \\
                \hline
                & MNLI & 4.9e-4 & Null & Null & Null & 100 \\
                & QNLI & 4.9e-4 & Null & Null & Null & 80 \\
                & CoLA & 4.9e-4 & Null & Null & Null & 5 \\
                & SST-2 & 4.9e-4 & Null & Null & Null & 50 \\
                \multirow{-5}{*}{LoRA-Over-MPO} & MRPC & 4.9e-4 & Null & Null & Null & 5 \\
                \hline
                & MNLI & 5e-4 & 4.9e-4 & Null & 9 & 100 \\
                & QNLI & 5e-4 & 4.9e-4 & Null & 9 & 80 \\
                & CoLA & 5e-4 & 4.9e-4 & Null & 9 & 5 \\
                & SST-2 & 5e-4 & 4.9e-4 & Null & 9 & 50 \\
                \multirow{-5}{*}{LoRA-Over-MPO$_P$} & MRPC & 5e-4 & 4.9e-4 & Null & 9 & 5 \\
                \hline
                & MNLI & 5e-4 & 4.9e-4 & 5 & 9 & 100 \\
                & QNLI & 5e-4 & 4.9e-4 & 7 & 9 & 80 \\
                & CoLA & 5e-4 & 4.9e-4 & 8 & 9 & 5 \\
                & SST-2 & 5e-4 & 4.9e-4 & 7 & 9 & 50 \\
                \multirow{-5}{*}{LoRA-Over-MPO$_R$} & MRPC & 5e-4 & 4.9e-4 & 8 & 9 & 5 \\
                \hline
                \end{tabular}
        }
    \label{tab:hyper_glue}
\end{table}

%% file: section/table/mpo_structure_T5_Base.tex
\begin{table}[htb!]
\caption{Summary of the MPO structure (T5-Base).}
    \centering
        \resizebox{.55\columnwidth}{!}{
            \begin{tabular}{c|c|c}
                \hline
                Shape & SVD & MPO \\
                \hline
                (768,8) & $\mathcal{T}^{24,32}_{2,4}(D)$ & $\mathcal{T}^{24,1,1,1,1,1,1,1,1,1,1,1,1,1,1,32}_{2,1,1,1,1,1,1,1,1,1,1,1,1,1,1,4}(D)$\\
                (8,768) & $\mathcal{T}^{2,4}_{24,32}(D)$ & $\mathcal{T}^{2,1,1,1,1,1,1,1,1,1,1,1,1,1,1,4}_{24,1,1,1,1,1,1,1,1,1,1,1,1,1,1,32}(D)$\\
                (3072,8) & $\mathcal{T}^{48,64}_{2,4}(D)$ & $\mathcal{T}^{48,1,1,1,1,1,1,1,1,1,1,1,1,1,1,64}_{2,1,1,1,1,1,1,1,1,1,1,1,1,1,1,4}(D)$\\
                (8,3072) & $\mathcal{T}^{2,4}_{48,64}(D)$ & $\mathcal{T}^{2,1,1,1,1,1,1,1,1,1,1,1,1,1,1,4}_{48,1,1,1,1,1,1,1,1,1,1,1,1,1,1,64}(D)$\\
                \hline
                \end{tabular}
    }
    
    \label{tab:mpo_structure_T5-base}
\end{table}

%% file: section/table/hyper_LLMs.tex
\begin{table}[htb!]
\large
\caption{Hyperparameter setup of LoRA-Over for Llama 2-7B and Llama 3.1-8B model. "LR" denote the learning rate. "Null" denote the parameter is useless.}
    \centering
        \resizebox{.99\columnwidth}{!}{
            \begin{tabular}{c|c|c|c|c|c|c|c}
                \hline
                 LLM & Method & Datasets & LR & MPO\_LR & split number & top-$N$ & eval step \\
                \hline
               & & MT-Bench & 9.3e-5 & Null & Null & Null & 50  \\
               & & GSM8K & 9.3e-5 & Null & Null & Null & 50 \\
               & \multirow{-3}{*}{LoRA-Over-SVD} & HumanEval & 1.1e-4 & Null & Null & Null & 50 \\
                
                \cline{2-8}
                & & MT-Bench & 9.3e-5 & Null & Null & Null & 50  \\
                & & GSM8K & 9.3e-5 & Null & Null & Null & 50 \\
                & \multirow{-3}{*}{LoRA-Over-MPO} & HumanEval & 1.1e-4 & Null & Null & Null & 50 \\
                
                \cline{2-8}
                & & MT-Bench & 1e-4 & 9.3e-5 & Null & 30 & 50 \\
                & & GSM8K & 1e-4 & 9.3e-5 & Null & 30 & 50 \\
                & \multirow{-3}{*}{LoRA-Over-MPO$_P$} & HumanEval & 1.2e-4 & 1.1e-4 & Null & 30 & 50 \\
                \cline{2-8}
                
                & & MT-Bench & 1e-4 & 9.3e-5 & 6 & 30 & 50  \\
                & & GSM8K & 1e-4 & 9.3e-5 & 8 & 30 & 50 \\
                \multirow{-12}{*}{Llama 2-7B~\cite{touvron2023llama}} & \multirow{-3}{*}{LoRA-Over-MPO$_R$} & HumanEval & 1.2e-4 & 1.1e-4 & 8 & 30 & 50 \\
                \hline
                
                & & MT-Bench & 1.7e-4 & Null & Null & Null & 50  \\
                & & GSM8K & 1.7e-4 & Null & Null & Null & 50 \\
                & \multirow{-3}{*}{LoRA-Over-SVD} & HumanEval & 7e-5 & Null & Null & Null & 50 \\
                \cline{2-8}
                
                & & MT-Bench & 1.7e-4 & Null & Null & Null & 50  \\
                & & GSM8K & 1.7e-4 & Null & Null & Null & 50 \\
                & \multirow{-3}{*}{LoRA-Over-MPO} & HumanEval & 7e-5 & Null & Null & Null & 50 \\
                \cline{2-8}
                
                & & MT-Bench & 2e-4 & 1.7e-4 & Null & 30 & 50 \\
                & & GSM8K & 2e-4 & 1.7e-4 & Null & 30 & 50 \\
                & \multirow{-3}{*}{LoRA-Over-MPO$_P$} & HumanEval & 1e-4 & 7e-5 & Null & 28 & 50 \\
                \cline{2-8}
            
                & & MT-Bench & 2e-4 & 1.7e-4 & 6 & 30 & 50  \\
                & & GSM8K & 2e-4 & 1.7e-4 & 8 & 30 & 50 \\
                \multirow{-12}{*}{Llama 3.1-8B~\cite{grattafiori2024llama}} & \multirow{-3}{*}{LoRA-Over-MPO$_R$} & HumanEval & 1e-4 & 7e-5 & 4 & 28 & 50 \\
                \hline
                \end{tabular}
        }
    
    \label{tab:hyper_LLMs}
\end{table}

%% file: section/table/mpo_structure_llama.tex
\begin{table*}[htb!]
\small
\caption{Summary of the MPO structure (Llama 2-7B and Llama 3.1-8B).}
    \centering
        \resizebox{.75\columnwidth}{!}{
            \begin{tabular}{c|c|c|c}
                \hline
                Shape & MT-Bench & GSM8K & HumanEval \\
                \hline
                \multicolumn{4}{c}{Llama 2-7B~\citep{touvron2023llama}} \\
                \hline
                \rowcolor{gray!10}
                \multicolumn{4}{c}{SVD}\\
                \hline
                (4096,8) & $\mathcal{T}^{64,64}_{2,4}(D)$ & $\mathcal{T}^{64,64}_{2,4}(D)$ & $\mathcal{T}^{64,64}_{2,4}(D)$ \\
                (8,4096) & $\mathcal{T}^{2,4}_{64,64}(D)$ & $\mathcal{T}^{2,4}_{64,64}(D)$ & $\mathcal{T}^{2,4}_{64,64}(D)$\\
                (11008,8) & $\mathcal{T}^{86,128}_{2,4}(D)$ & $\mathcal{T}^{86,128}_{2,4}(D)$ & $\mathcal{T}^{86,128}_{2,4}(D)$ \\
                (8,11008) & $\mathcal{T}^{2,4}_{86,128}(D)$ & $\mathcal{T}^{2,4}_{86,128}(D)$ & $\mathcal{T}^{2,4}_{86,128}(D)$ \\
                \hline
                \rowcolor{gray!10}
                \multicolumn{4}{c}{MPO}\\
                \hline
                (4096,8) & $\mathcal{T}^{64,1,1,1,1,1,1,1,64}_{2,1,1,1,1,1,1,1,4}(D)$ & $\mathcal{T}^{64,1,1,1,1,1,1,1,64}_{2,1,1,1,1,1,1,1,4}(D)$ & $\mathcal{T}^{64,1,1,1,1,1,1,1,64}_{2,1,1,1,1,1,1,1,4}(D)$ \\
                (8,4096) & $\mathcal{T}^{2,1,1,1,1,1,1,1,4}_{64,1,1,1,1,1,1,1,64}(D)$ & $\mathcal{T}^{2,1,1,1,1,1,1,1,4}_{64,1,1,1,1,1,1,1,64}(D)$ & $\mathcal{T}^{2,1,1,1,1,1,1,1,4}_{64,1,1,1,1,1,1,1,64}(D)$\\
                (11008,8) & $\mathcal{T}^{86,1,1,1,1,1,1,1,128}_{2,1,1,1,1,1,1,1,4}(D)$ & $\mathcal{T}^{86,1,1,1,1,1,1,1,128}_{2,1,1,1,1,1,1,1,4}(D)$ & $\mathcal{T}^{86,1,1,1,1,1,1,1,128}_{2,1,1,1,1,1,1,1,4}(D)$ \\
                (8,11008) & $\mathcal{T}^{2,1,1,1,1,1,1,1,4}_{86,1,1,1,1,1,1,1,128}(D)$ & $\mathcal{T}^{2,1,1,1,1,1,1,1,4}_{86,1,1,1,1,1,1,1,128}(D)$ & $\mathcal{T}^{2,1,1,1,1,1,1,1,4}_{86,1,1,1,1,1,1,1,128}(D)$ \\
                \hline
                \multicolumn{4}{c}{Llama 3.1-8B~\citep{grattafiori2024llama}} \\
                \hline
                \rowcolor{gray!10}
                \multicolumn{4}{c}{SVD}\\
                \hline
                (4096,8) & $\mathcal{T}^{64,64}_{2,4}(D)$ & $\mathcal{T}^{64,64}_{2,4}(D)$ & $\mathcal{T}^{64,64}_{2,4}(D)$ \\
                (8,4096) & $\mathcal{T}^{2,4}_{64,64}(D)$ & $\mathcal{T}^{2,4}_{64,64}(D)$ & $\mathcal{T}^{2,4}_{64,64}(D)$\\
                (1024,8) & $\mathcal{T}^{32,32}_{2,4}(D)$ & $\mathcal{T}^{32,32}_{2,4}(D)$ & $\mathcal{T}^{32,32}_{2,4}(D)$ \\
                (8,1024) & $\mathcal{T}^{2,4}_{32,32}(D)$ & $\mathcal{T}^{2,4}_{32,32}(D)$ & $\mathcal{T}^{2,4}_{32,32}(D)$\\
                (14336,8) & $\mathcal{T}^{112,128}_{2,4}(D)$ & $\mathcal{T}^{112,128}_{2,4}(D)$ & $\mathcal{T}^{112,128}_{2,4}(D)$ \\
                (8,14336) & $\mathcal{T}^{2,4}_{112,128}(D)$ & $\mathcal{T}^{2,4}_{112,128}(D)$ & $\mathcal{T}^{2,4}_{112,128}(D)$ \\
                \hline
                \rowcolor{gray!10}
                \multicolumn{4}{c}{MPO}\\
                \hline
                (4096,8) & $\mathcal{T}^{64,1,1,1,1,1,1,1,1,64}_{2,1,1,1,1,1,1,1,1,4}(D)$ & $\mathcal{T}^{64,1,1,1,1,1,1,1,1,64}_{2,1,1,1,1,1,1,1,1,4}(D)$ & $\mathcal{T}^{64,1,1,1,1,1,1,64}_{2,1,1,1,1,1,1,4}(D)$ \\
                (8,4096) & $\mathcal{T}^{2,1,1,1,1,1,1,1,1,4}_{64,1,1,1,1,1,1,1,1,64}(D)$ & $\mathcal{T}^{2,1,1,1,1,1,1,1,1,4}_{64,1,1,1,1,1,1,1,1,64}(D)$ & $\mathcal{T}^{2,1,1,1,1,1,1,4}_{64,1,1,1,1,1,1,64}(D)$\\
                (1024,8) & $\mathcal{T}^{32,1,1,1,1,1,1,1,1,32}_{2,1,1,1,1,1,1,1,1,4}(D)$ & $\mathcal{T}^{32,1,1,1,1,1,1,1,1,32}_{2,1,1,1,1,1,1,1,1,4}(D)$ & $\mathcal{T}^{32,1,1,1,1,1,1,32}_{2,1,1,1,1,1,1,4}(D)$ \\
                (8,1024) & $\mathcal{T}^{2,1,1,1,1,1,1,1,1,4}_{32,1,1,1,1,1,1,1,1,32}(D)$ & $\mathcal{T}^{2,1,1,1,1,1,1,1,1,4}_{32,1,1,1,1,1,1,1,1,32}(D)$ & $\mathcal{T}^{2,1,1,1,1,1,1,4}_{32,1,1,1,1,1,1,32}(D)$\\
                (14336,8) & $\mathcal{T}^{112,1,1,1,1,1,1,1,1,128}_{2,1,1,1,1,1,1,1,1,4}(D)$ & $\mathcal{T}^{112,1,1,1,1,1,1,1,1,128}_{2,1,1,1,1,1,1,1,1,4}(D)$ & $\mathcal{T}^{112,1,1,1,1,1,1,128}_{2,1,1,1,1,1,1,4}(D)$ \\
                (8,14336) & $\mathcal{T}^{2,1,1,1,1,1,1,1,1,4}_{112,1,1,1,1,1,1,1,1,128}(D)$ & $\mathcal{T}^{2,1,1,1,1,1,1,1,1,4}_{112,1,1,1,1,1,1,1,1,128}(D)$ & $\mathcal{T}^{2,1,1,1,1,1,1,4}_{112,1,1,1,1,1,1,128}(D)$ \\
                \hline
                \end{tabular}
    }

    \label{tab:mpo_structure_llama}
\end{table*}

%% file: section/table/memory.tex
\begin{table}
\small
\caption{Comparison of training time and peak
GPU memory on Llama 2-7B and Llama 3.1-8B.}
    \centering
        \resizebox{.5\columnwidth}{!}{
            \begin{tabular}{l|cc}
            \toprule
                \textbf{Method} & \textbf{Training Time} & \textbf{Peak Memory} \\
                \midrule
                \rowcolor{gray!20}\multicolumn{3}{c}{\textbf{Llama 2-7B~\citep{touvron2023llama}}}    \\
                \midrule
                LoRA & 3h47min & 42.74GB \\
                LoRA-Over-MPO$_R$ & 19h25min & 17.66GB \\
                \midrule
                \rowcolor{gray!20}\multicolumn{3}{c}{\textbf{Llama 3.1-8B~\citep{grattafiori2024llama}}}    \\
                \midrule
                LoRA & 4h05min & 49.66GB \\
                LoRA-Over-MPO$_R$ & 16h34min & 19.94GB \\
                \bottomrule
                \end{tabular}
        }
    \label{tab:memory}
\end{table}

%% file: section/table/parameter_count.tex
\begin{table*}[htb!]
\small
\caption{Comparison of trainable parameters count. "\# To (M)-Train" and "\#
To (M)-Inference" denote the number (in millions) of total parameters during training and test, respectively.}
    \centering
        \resizebox{.8\columnwidth}{!}{
            \begin{tabular}{l|cccc}
            \toprule
                \textbf{Method} & \textbf{Dataset} & \textbf{Target Modules} & \textbf{\#To (M)-Train} & \textbf{\#To (M)-Inference} \\
                \midrule
                \rowcolor{gray!20}\multicolumn{5}{c}{\textbf{T5-Base~\citep{raffel2020exploring}}}    \\
                \midrule
                LoRA & SST-2 & all-linear & 3.24M & 3.24M \\
                LoRA-Over-MPO$_R$ & SST-2 & all-linear & 16.93M & 3.24M \\
                \midrule
                \rowcolor{gray!20}\multicolumn{5}{c}{\textbf{Llama 2-7B~\citep{touvron2023llama}}}    \\
                \midrule
                LoRA & MetamathQA & all-linear & 19.99M & 19.99M \\
                LoRA-Over-MPO$_R$ & MetamathQA & all-linear & 84.54M & 19.99M \\
                \midrule
                \rowcolor{gray!20}\multicolumn{5}{c}{\textbf{Llama 3.1-8B~\citep{grattafiori2024llama}}}    \\
                \midrule
                LoRA & MetamathQA & attention  & 6.82M & 6.82M \\
                LoRA-Over-MPO$_R$ & MetamathQA & attention & 35.57M & 6.82M \\
                \bottomrule
                \end{tabular}
        }
    
    \label{tab:parameter_count}
\end{table*}